\documentclass[letterpaper, 10 pt, conference]{ieeeconf}
\IEEEoverridecommandlockouts
\usepackage{graphicx}
\usepackage{amsmath}
\usepackage{amssymb}
\usepackage{booktabs}
\usepackage{multirow}
\usepackage{cite}
\usepackage{stfloats} 
\usepackage{subcaption}
\usepackage{algorithm}
\usepackage{algpseudocode}
\usepackage{makecell}

\makeatletter
\newcommand\fs@spacedruled{%
  \def\@fs@cfont{\bfseries}%
  \let\@fs@capt\floatc@ruled
  \def\@fs@pre{%
    \kern4pt
    \hrule height .8pt depth 0pt
    \kern2pt
  }%
  \def\@fs@post{%
    \kern2pt
    \hrule\relax
  }%
  \def\@fs@mid{%
    \kern2pt
    \hrule
    \kern2pt
  }%
  \let\@fs@iftopcapt\iftrue
}
\floatstyle{spacedruled}
\restylefloat{algorithm}
\makeatother

\usepackage{xspace}
\usepackage[dvipsnames,table,xcdraw]{xcolor}

\newcommand{\Method}[1]{\textup{#1}}
\definecolor{almostblack}{rgb}{0, 0, 0.3}

\DeclareRobustCommand{\MRMP}{\Method{MRMP}\xspace}
\DeclareRobustCommand{\ARC}{\Method{ARC}\xspace}
\DeclareRobustCommand{\PRM}{\Method{PRM}\xspace}

\DeclareRobustCommand{\RRTC}{\Method{RRT-C}\xspace}

\DeclareRobustCommand{\SPITE}{\Method{SPITE}\xspace}
\DeclareRobustCommand{\MRSPITE}{\Method{MR-SPITE}\xspace}
\DeclareRobustCommand{\MRSPITEH}{\Method{MR-SPITE-H}\xspace}
\DeclareRobustCommand{\VAMP}{\Method{VAMP}\xspace}

\DeclareRobustCommand{\VAMPH}{\Method{VAMP-H}\xspace}
\DeclareRobustCommand{\dRRT}{\Method{dRRT}\xspace}
\DeclareRobustCommand{\DRRTstar}{\Method{dRRT*}\xspace}

\DeclareRobustCommand{\FDRRTstar}{\Method{Fast-dRRT*}\xspace}
\DeclareRobustCommand{\RRTC}{\Method{RRT-C}\xspace}
\DeclareRobustCommand{\FFC}{\Method{FFC}\xspace}
\DeclareRobustCommand{\AABB}{\Method{AABB}\xspace}
\DeclareRobustCommand{\OBB}{\Method{OBB}\xspace}

\title{\LARGE \bf
MR-SPITE: Accelerating Multi-Robot Conflict Scans via Hierarchical Swept-Volume Approximations
}

\author{Marta Markowicz$^{1}$, James Motes$^{1}$, Marco Morales$^{1, 2}$, and Nancy M. Amato$^{1}$%
\thanks{$^{1}$The authors are with the Siebel School of Computing and Data Science, University of Illinois, 201 N. Goodwin Avenue, Urbana, IL 61801, USA.}%
\thanks{$^{2}$Marco Morales is also affiliated with the Department of Computer Science at Instituto Tecnol\'ogico Aut\'onomo de M\'exico (ITAM), Mexico City, M\'exico.}%
}

\begin{document}

\maketitle
\vspace{-8mm}
\thispagestyle{empty}
\pagestyle{empty}

\begin{abstract}
Conflict scanning over synchronized robot paths requires detailed collision checking, potentially across every robot pair at every timestep, and may be repeated many times as conflicts are repaired.
We present Multi-Robot SPITE (\MRSPITE), a conservative, motion-segment-based filter for accelerating these scans. \MRSPITE partitions each path into temporal intervals and assigns conservative bounds to each segment. An interval scheduler compares bounds for temporally overlapping motions: disjoint bounds certify the shared window as conflict-free, while unresolved windows are passed to the underlying collision checker. We integrate \MRSPITE into \ARC and combine it with VAMP-based collision checking.
For 16 Fetch robots, \ARC with \MRSPITE achieves a
paired median conflict scan speedup of $7.18\times$ and reduces median
planning time by 57\% relative to the baseline \ARC implementation with \PRM+\VAMP.
These results demonstrate that motion-segment bounds complement configuration-level collision acceleration while preserving the behavior of the underlying discretized scanner.

\end{abstract}


\section{Introduction}

Multi-robot systems enable concurrent operation in applications such as warehouse automation and industrial manufacturing.
Realizing these benefits requires multi-robot motion planning (\MRMP) to generate collision-free motions through shared workspaces.
Scaling \MRMP presents two distinct computational challenges: the combinations of individual robot states grow exponentially with the number of robots, while inter-robot validation faces a quadratic growth in the number of potentially interacting robot pairs.

Conflict-driven methods address the search-space challenge by initially planning for robots independently and resolving interactions as conflicts are discovered~\cite{sharon2015_cbs,solis2021_cbsmp,sim2025_stcbs,solis-arc-24}.
These methods lazily validate candidate paths and use detected conflicts to guide replanning.
For methods that resolve the earliest collision, this \textit{find-first-conflict} (FFC) operation is repeated as paths are revised.

Recent work~\cite{motes2026_va_mrmp,huang2026_vamp_mr} extends the vectorized validation of single robot planning~\cite{vamp_2024} to accelerate multi-robot collision checking.
Nevertheless, repeated conflict scans can remain a substantial portion of planning time.
This motivates a complementary approach: eliminate groups of detailed collision checks rather than solely accelerating their execution.

In this paper, we present Multi-Robot SPITE (\MRSPITE), a hierarchical swept-volume approximation technique for accelerating FFC.
The original SPITE method~\cite{ashur-spite-24} uses conservative bounds on roadmap motions to rapidly revalidate them after obstacle changes.
MR-SPITE extends this approach to synchronized robot paths, using the geometry and timing of motion segments to rule out conflicts over entire intervals.

We evaluate \MRSPITE within \ARC~\cite{solis-arc-24}, a conflict-driven \MRMP approach, on mobile robots and articulated manipulators.
For sixteen Fetch robots, \MRSPITE achieves a 7.2$\times$ conflict scan speedup and more than halves median planning time relative to the baseline \ARC configuration.
These results demonstrate that motion-level filtering can complement vectorized collision checking of configurations to improve overall planning performance.
In summary, our contributions are:
\begin{itemize}
    \item A conservative motion-interval filter that eliminates detailed collision checks over portions of synchronized multi-robot paths.
    \item A time-aware, hierarchical conflict scanner that preserves the underlying discretized scanner’s earliest-conflict result.
    \item An evaluation of scan and planning-time improvements across robot types, team sizes, and collision-model complexities.
\end{itemize}

\begin{figure*}
    \centering
    \vspace{5mm} 
    \includegraphics[width=0.7\linewidth,page=1]{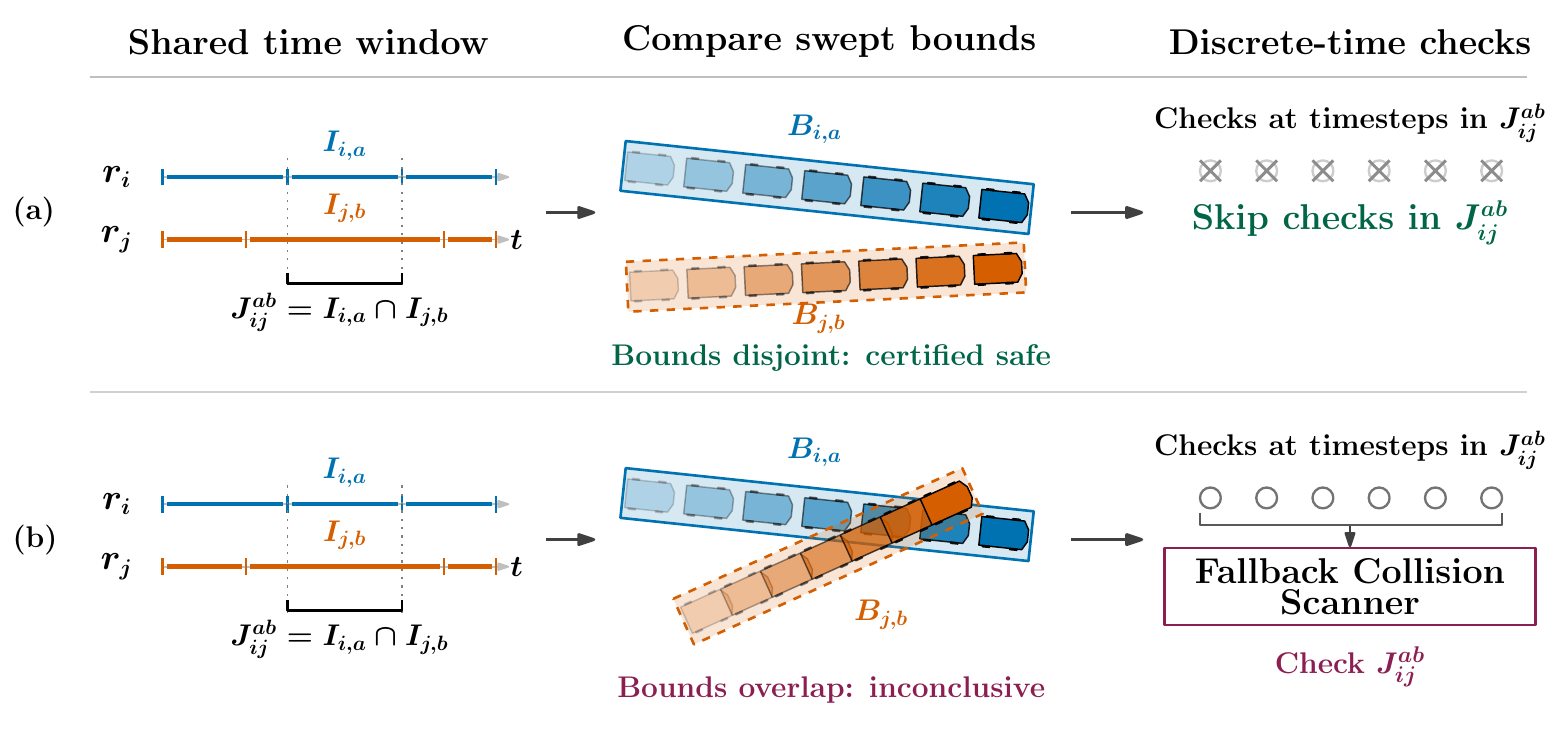}
    \vspace{-2mm} 
    \caption{ \small\textbf{Collision Scanner Filtering}
    \MRSPITE compares conservative swept bounds for temporally overlapping path intervals. (a) Disjoint bounds certify the corresponding discrete-time window as conflict-free, allowing all detailed checks in that window to be skipped. (b) Overlapping bounds are inconclusive and invoke the underlying collision checker.
    }
    \label{fig:mrspite_interval_filtering}
    \vspace{-6mm} 
\end{figure*}

\section{Related Work}

This section reviews related work in (A) multi-robot motion planning, (B) methods for collision checking and how to accelerate it, and (C) swept-volume techniques for motion validation.

\subsection{Multi-Robot Planning}

\MRMP can be formulated as a search directly in the composite 
configuration space formed by the Cartesian product of the individual 
robots' configuration spaces.
Explicitly constructing this space is generally
impractical, so roadmap-based methods such as \dRRT~\cite{solovey-mrdrrt-16} and \DRRTstar~\cite{shome2020_drrtstar} represent this space implicitly with combinations of vertices and edges from individual-robot roadmaps. Even with this simplified representation, the number of possible joint states and transitions to explore grows rapidly with the number of robots.

Other approaches construct paths in the individual configuration spaces. Prioritized planning plans robots sequentially in a priority ordering, treating the trajectories of higher-priority robots as moving obstacles when planning for lower-priority robots~\cite{van2005_prioritized}. Prioritized Safe-Interval Path Planning with Continuous-Time Conflicts
(PSIPP/CTC) extends prioritized planning to continuous-time motion on
2D roadmaps. It precomputes conflict intervals between roadmap
vertices and edges and uses them to identify safe time intervals during
sequential planning~\cite{kasaura2022_psipp}.

Conflict-driven methods construct a set of individual paths and then identify and resolve conflicts. Conflict-Based Search (CBS) resolves conflicts using a tree of constraints on individual agents~\cite{sharon2015_cbs}. CBS-MP adapts this strategy to continuous multi-robot motion planning by searching individual robot roadmaps~\cite{solis2021_cbsmp}, while ST-CBS combines high-level conflict resolution with low-level planning in continuous space-time~\cite{sim2025_stcbs}.
\ARC uses detected conflicts to construct local coupled subproblems and repair the affected paths~\cite{solis-arc-24}. 

Many conflict-driven methods must repeatedly scan 
synchronized paths to identify inter-robot conflicts.  
For $N$ robots and $K$ synchronized timesteps, a complete scan
may evaluate as many as $K\binom{N}{2}$ robot-pair/configuration
queries. Because conflict-driven planners can invoke this operation
after every repair, conflict scanning may become a substantial
repeated cost even when planning is performed primarily in the
individual configuration spaces.

\subsection{Collision-Checking Acceleration}

Collision checking is often a major computational cost in both single- and multi-robot motion planning~\cite{elbanhawi-sbmpRev-14}. Motion validity is commonly approximated by sampling configurations along a trajectory and checking each sample for collision, requiring numerous forward-kinematics evaluations and intersection tests at each configuration.
Existing methods to accelerate collision detection can be broadly grouped into four categories: sampling strategies, hierarchical filtering,
simplified collision geometry, and hardware-accelerated execution.

Sampling strategies can reduce or reorder the configurations evaluated along a motion. Typical uniform distribution checks configurations at a fixed discretization, whereas
adaptive dynamic collision checking varies the sampling
resolution along a path, adding more samples in areas with higher likelihood of collision~\cite{schwarzer2005_adaptive_cc}. Bisection strategies prioritize samples that are likely to lead to early termination~\cite{sanchez2003_bidirectional_prm}. Similarly, \VAMP introduces a SIMD-based rake that evaluates distributed samples along a motion in parallel, increasing the likelihood of early termination for invalid motions~\cite{vamp_2024}.

Hierarchical filtering uses cheap conservative tests before performing more detailed collision checks. Bounding-volume hierarchies bound individual configurations to narrow down detailed geometric tests to specific sections of the robot~\cite{gottschalk1996_obbtree}. 
Motion-level hierarchies bound the volume of an entire trajectory segment, allowing filtering of a whole set of configurations. This is further discussed in Section~\ref{sec:swept_volumes}.

Simplified collision models approximate complex geometry using sets or hierarchies of primitives such as spheres, capsules, or boxes, replacing many triangle-intersection tests from meshes~\cite{pan2012fcl}. Sphere-based models require only center-distance comparisons, significantly reducing the collision checking cost~\cite{hubbard-spheres-96}. They  have a tradeoff, however, between model size and geometric tightness, where coarse models are inexpensive but overly conservative, and detailed models require more primitive pair tests.

Hardware-accelerated methods utilize parallelism and vectorization. \VAMP combines struct-of-arrays data layouts, vectorized forward 
kinematics, and SIMD sphere-based collision checking to validate
batches of configurations efficiently~\cite{vamp_2024}. Related multi-robot works apply similar ideas of SIMD-accelerated collision checking and vectorization to inter-robot validation~\cite{huang2026_vamp_mr,motes2026_va_mrmp}.

\subsection{Swept-Volume and Motion-Interval Filtering}\label{sec:swept_volumes}

The swept volume of a motion segment is the union of the workspace occupied by the robot during the motion. Exact computation of this volume is expensive, so collision-checking methods use conservative approximations such as bounding boxes, hierarchical bounds, or voxel sets.
Motion Oriented Bounding Box trees use
hierarchical bounds for characters in reusable animation clips
\cite{sung2005_mobb}. \FDRRTstar precomputes voxelized swept volumes for roadmap transitions in multi-robot planning~\cite{solano2023_fastdrrtstar}.
Such conservative bounds can be compared against other geometry; when the bounds are disjoint, all underlying configuration-level collision checks can be omitted.

\SPITE precomputes conservative swept-volume approximations for the vertices and edges of a probabilistic roadmap~\cite{ashur-spite-24}, specifically line-swept-spheres or alternatively Oriented Bounding Boxes (OBBs). For articulated robots, each link's swept volume is approximated by a separate swept-volume approximation, allowing a tighter bound than one approximation for the whole robot. At runtime, the approximations allow motions to be quickly revalidated in response to changes in the environment, allowing for lazy collision detection where only edges with intersected bounds need to be checked by the expensive fallback collision checker.  

The initial paths in many conflict-driven planners can be generated with roadmaps and consist mostly of a sequence of PRM edges. \MRSPITE partitions the synchronized paths into temporal intervals corresponding to roadmap edges, query connectors, goal holds, and locally repaired motions. It associates each interval with a conservative \SPITE-style bound and schedules only interval pairs whose time ranges overlap. Disjoint bounds certify the entire shared time range as conflict-free; overlapping bounds are inconclusive and are passed to an existing configuration-level scanner such as \VAMP. \MRSPITE thus enhances collision checking in conflict-driven planners by extending swept-volume filtering to the collision scan.

\section{Problem Definition}

Consider a set of $N$ robots
$\mathcal{R}={r_1,\ldots,r_N}$ operating in a shared workspace.
Each robot $r_i$ is assigned a time-parameterized path
$\pi_i:[0,T]\rightarrow\mathcal{C}_i$, where $\mathcal{C}_i$ is its
configuration space. Let $\mathcal{S}_i(q)\subseteq\mathbb{R}^d$ denote
the workspace occupied by $r_i$ at configuration $q$.

Two robots $r_i$ and $r_j$ conflict at time $t$ if
\begin{equation}
\mathcal{S}_i(\pi_i(t))\cap
\mathcal{S}_j(\pi_j(t))\neq\emptyset.
\end{equation}
The paths are conflict-free if this intersection is empty for every
$t\in[0,T]$.

In practice, paths are represented by synchronized sequences of
configurations evaluated at discrete times
$\mathcal{T}={\tau_0,\ldots,\tau_{K-1}}$. We write
$q_{i,k}=\pi_i(\tau_k)$ for the configuration of robot $r_i$ at
timestep $k$. Paths that terminate early are extended by holding their
final configurations so that every robot has a configuration at each
synchronized timestep. Let
\begin{equation}
\operatorname{Collide}(i,j,k)
=
[
\mathcal{S}_i(q_{i,k})
\cap
\mathcal{S}_j(q_{j,k})
\neq \emptyset] .
\end{equation}
denote the result of the underlying configuration-level collision
checker.

Many multi-robot planners first compute paths in the individual robots'
configuration spaces and then use detected inter-robot conflicts to guide
search or path repair. We refer to this class as
\emph{conflict-driven} methods. Given the synchronized paths
$\Pi=\{\pi_1,\ldots,\pi_N\}$, the conflict scanner examines timesteps
in increasing order. Let $k^*$ be the first timestep at which any pair
of robots collides. The scanner returns $(k^*,i^*,j^*)$, where
$(i^*,j^*)$ is the first colliding pair at timestep $k^*$ under a fixed
deterministic pair ordering. If no conflict exists, it returns $\varnothing$.
A conflict-driven planner may repeat this scan whenever its paths are
modified or repaired.

\section{Method}

In this section, we present \MRSPITE, a method for accelerating the
\FFC operation in roadmap-based, conflict-driven \MRMP planners. During preprocessing,
we construct an individual roadmap for each robot together with \SPITE-style
conservative swept-volume bounds for its vertices and edges.
These roadmaps and geometries can be reused across many \MRMP queries.
At query time, the chosen planner obtains an initial path for each robot from
its corresponding roadmap. \MRSPITE is then used to accelerate
conflict scanning between the synchronized paths through lazy collision checking.

The method is described in four sections. We first describe how
the path is partitioned into intervals and how the conservative bounds
are sourced (Sec.~\ref{sec:path_inter}). We then present the
interval scheduler used to process robot pairs in temporal order (Sec.~\ref{sec:scanner}),
followed by details on the fallback collision checking and optional per-link filtering.
Finally, we describe how \MRSPITE is integrated into the
\ARC planner and how interval annotations are maintained when paths are
repaired or synchronized.

\subsection{Path Intervals and Bounds}\label{sec:path_inter}

For each robot $r_i$, \MRSPITE partitions the discrete path domain into
an ordered collection of half-open intervals
\begin{equation}
\mathcal{I}_i =
{I_{i,1},\ldots,I_{i,m_i}},
\qquad
I_{i,a}=[b_{i,a},e_{i,a}),
\end{equation}
such that the intervals are non-overlapping and cover
$[0,K)$. A timestep $k$ belongs to exactly one interval of each robot.

In our implementation, an interval may
correspond to a precomputed PRM edge, a start or goal query connector,
a goal-hold segment, or an online local repair. The source of the motion interval determines how the interval's conservative geometry is
obtained and allows interval annotations to be preserved when paths
are padded or modified.


For intervals corresponding
to PRM edges, the \SPITE method is used to compute whole-robot and per-link conservative OBBs offline from the collision geometry at densely sampled configurations along each
edge. These bounds are stored with the roadmap and retrieved as needed by the motion intervals.
For motion intervals created online, including connections from the start and goal to the roadmap, goal waiting behavior, and local repairs, \MRSPITE constructs supplemental bounds during runtime. Whole-body and per-link \AABB{}s are constructed by computing the maximum extents of the spheres at every intermediate configuration along a motion.
Thus for every motion interval along a robot's trajectory we have an assigned bound 
that conservatively encloses the corresponding discretized collision model.


Consider intervals $I_{i,a}$ and $I_{j,b}$ for robots $r_i$ and $r_j$.
Only the timesteps in their shared temporal window
\begin{equation}
J_{ij}^{ab}
=
I_{i,a}\cap I_{j,b}
=
[\max(b_{i,a},b_{j,b}),
\min(e_{i,a},e_{j,b}))
\end{equation}
can be evaluated together.

Figure~\ref{fig:mrspite_interval_filtering} illustrates the two
possible outcomes. The left column identifies the shared window
$J_{ij}^{ab}$, the center column compares the intervals' swept bounds,
and the right column shows the resulting collision-checking work. In
Fig.~\ref{fig:mrspite_interval_filtering}(a), the bounds are disjoint, and thus with a single bound comparison we have certified every sampled
configuration pair in $J_{ij}^{ab}$ as safe.
In Fig.~\ref{fig:mrspite_interval_filtering}(b), the bounds overlap, but this may be a result of the conservatism of the bounds. Consequently, the
configuration pairs in $J_{ij}^{ab}$ are submitted to the underlying
collision checker.

\subsection{Interval-Scheduled Conflict Scanning}\label{sec:scanner}

\begin{algorithm}[t]
\caption{MR-SPITE interval-scheduled conflict scan}
\label{alg:mr-spite-scheduler}
\small
\begin{algorithmic}[1]
\Require Paths $\Pi=\{\pi_1,\ldots,\pi_N\}$ with $K$ timesteps,
interval partitions $\{\mathcal{I}_i\}$, and bound hierarchies
$\{\mathcal{B}_i\}$
\Ensure Earliest conflict $(k,i,j)$, or $\varnothing$

\State $Q\gets$ empty min-priority queue ordered by $(t,i,j)$
\State $C^*\gets\varnothing$
    \Comment{Earliest conflict found so far}

\ForAll{$1\leq i<j\leq N$}
    \State $\Call{Push}{Q,(0,i,j)}$
\EndFor

\While{$Q$ is not empty}
    \State $(t,i,j)\gets\Call{PopMin}{Q}$

    \If{$C^*\neq\varnothing$ \textbf{and}
        $t>\operatorname{time}(C^*)$}
        \State \textbf{break}
    \EndIf

    \State $I_{i,a}\gets\Call{IntervalAt}{\mathcal{I}_i,t}$
    \State $I_{j,b}\gets\Call{IntervalAt}{\mathcal{I}_j,t}$
    \State $e\gets\min(e_{i,a},e_{j,b})$

    \If{$C^*\neq\varnothing$}
        \State $e\gets\min(e,\operatorname{time}(C^*)+1)$
    \EndIf

    \If{$B_{i,a}\cap B_{j,b}=\emptyset$}
        \State $t'\gets e$
        \Comment{Whole window is conflict-free}
    \Else
        \State $L\gets
        \Call{CandidateLinkPairs}
        {\mathcal{B}_{i,a},\mathcal{B}_{j,b}}$

        \If{$L=\emptyset$}
            \State $t'\gets e$
            \Comment{No swept link pair can collide}
        \Else
            \State $k'\gets
            \Call{FindFirstConflict}
            {\pi_i,\pi_j,[t,e),L}$

            \If{$k'\neq\varnothing$}
                \State $C^*\gets
                \Call{LexMin}{C^*,(k',i,j)}$
                \State \textbf{continue}
            \EndIf

            \State $t'\gets e$
        \EndIf
    \EndIf

    \If{$t'<K$ \textbf{and}
        ($C^*=\varnothing$ \textbf{or}
        $t'\leq\operatorname{time}(C^*)$)}
        \State $\Call{Push}{Q,(t',i,j)}$
    \EndIf
\EndWhile

\State \Return $C^*$
\end{algorithmic}
\end{algorithm}

\MRSPITE maintains an independent cursor for every unordered robot
pair. As shown in Lines~1--5 of Algorithm~\ref{alg:mr-spite-scheduler},
the cursors are stored in a priority queue ordered by timestep and
robot-pair index. The variable $C^*$ records the earliest conflict
found so far under the baseline scanner's deterministic ordering.

At each iteration, the scheduler removes the pair with the earliest
unresolved timestep (Line~7). Lines~11--13 identify the intervals
containing that timestep and set $e$ to the end of their shared
window. If a candidate conflict has already been found, Lines~14--16
limit further checking to timesteps that could precede or tie that
candidate.

Lines~17--31 process the shared window. If the interval bounds are
disjoint, the complete window is certified conflict-free and the
cursor advances directly to $e$. Otherwise, the fallback scanner
returns the first conflict in the window. Line~26 updates $C^*$
using lexicographic ordering over $(k,i,j)$. If no conflict is found,
Lines~32--34 reinsert the pair at the next unresolved timestep.

The scheduler does not return immediately after finding a conflict.
Lines~8--9 terminate only when the earliest remaining cursor is later
than $\operatorname{time}(C^*)$. Therefore, before returning
$C^*$, every robot pair with unresolved work at that timestep or
earlier has been either certified by conservative bounds or evaluated
by the fallback scanner. The strict inequality in Line~8 also ensures
that equal-time conflicts are processed before deterministic
robot-pair tie breaking is finalized. Consequently, \MRSPITE returns
the same canonical earliest conflict as the baseline FFC scanner.

For each interval window, \MRSPITE performs a whole-robot bound test
followed, when necessary, by per-link OBB comparisons. Detailed
configuration checking is required only if at least one link pair
remains unresolved. In the worst case, all link bounds overlap and
\MRSPITE performs the baseline collision-checking work in addition to
its filtering overhead. When the swept link bounds are disjoint,
however, these inexpensive bound comparisons replace all
configuration-pair checks in the window.

\subsection{Hierarchical Filtering and Detailed Validation}

\MRSPITE first compares the whole-robot swept bounds for the current
interval window in Line~17. If these bounds are disjoint,
the complete window is certified conflict-free and its configuration
checks are skipped.
Line~20 calls \textsc{CandidateLinkPairs}, which first compares the
per-link swept \OBB{}s and identifies the pairs whose bounds overlap. If
none overlap, it returns the empty set and the complete interval is
certified conflict-free. If at least one pair overlaps, standard
\MRSPITE returns all robot-link pairs for the dense fallback, whereas
MR-SPITE-H returns only the overlapping pairs.

Line~24 passes the surviving link pairs to the configuration-level
fallback scanner; in this paper we specifically use \VAMP or hierarchical \VAMP (\VAMPH) as our 
fallback collision detector.
With standard \VAMP, $L$ determines which detailed sphere pairs are evaluated.
With hierarchical \VAMP (\VAMPH), the retained link pairs determine
which pairs of VAMP-H sphere groups remain candidates. Each group
contains the detailed collision spheres associated with a robot link.
For every retained group pair, VAMP-H first compares the groups'
enclosing spheres at the current configuration and evaluates their
detailed sphere pairs only if the enclosing spheres overlap.

These filters operate at different scales: \MRSPITE filters complete
temporal windows, \MRSPITEH filters swept link pairs over those
windows, and \VAMPH filters groups of spheres at individual
configurations. Combining MR-SPITE-H with \VAMPH introduces additional
mask-construction and restricted-fallback overhead, and the two
hierarchies may reject many of the same interactions. The combined
method is therefore beneficial only when the detailed work removed by
the link mask exceeds this additional overhead.

\subsection{Integration with \ARC}
\label{sec:arc-integration}

We integrate \MRSPITE into \ARC's existing find-first-conflict routine. 
\ARC generates initial paths for each robot from a roadmap.
These individual paths are annotated as start connectors, \PRM edges,
goal connectors, or goal holds. When \ARC replaces a path segment with a local repair, \MRSPITE constructs an \AABB for the repaired segment while preserving the source annotations of unaffected segments and updating their timestep indices. When paths are synchronized
by padding their endpoints with a wait, the appended interval is explicitly marked as a goal hold and an \AABB is constructed for the hold.

After every modification, \ARC invokes the same \FFC interface.
\MRSPITE changes only how that scan is evaluated; it does not change
which conflict is selected or how \ARC constructs a repair. Therefore,
given identical synchronized paths and the same underlying collision
checker, \ARC receives the same earliest conflict from the baseline and
\MRSPITE scanners.

\begin{figure}
    \vspace*{8pt}
    \centering
    \includegraphics[width=0.4\linewidth]{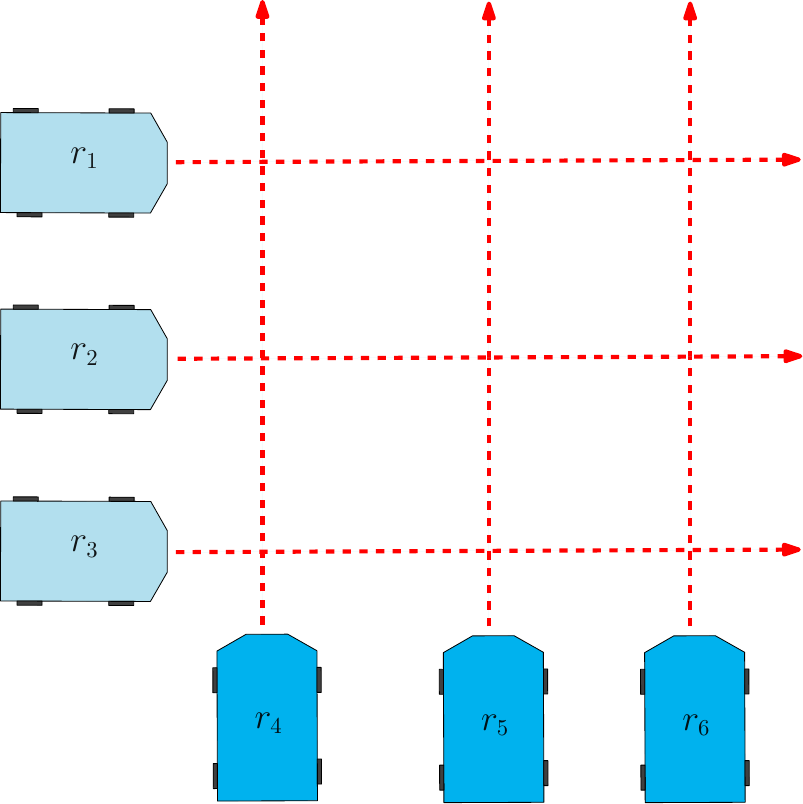}
    \caption{\small \textbf{Mobile crossing scene.} Robots follow perpendicular lanes
through a shared central crossing. The lane positions are randomly
jittered for each task while
preserving the crossing structure.}
    \label{fig:mobile_cross}

\end{figure}

\begin{figure}
    \centering
    \includegraphics[width=0.85\linewidth,page=4]{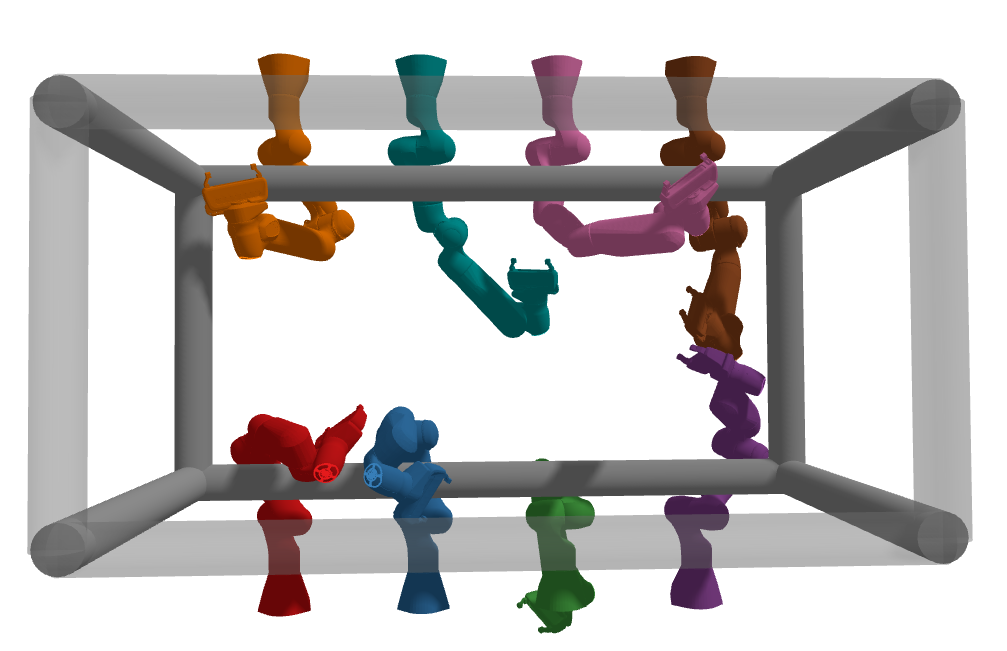}
    \caption{ \small \textbf{Cage with Franka Emika Panda manipulators.} Robots are randomly assigned valid query points inside or outside the cage.}
    \vspace{-4mm} 
    \label{fig:panda_cage}
\end{figure}
\vspace{-5mm}

\section{Evaluation}

Our evaluation addresses three questions: (1) how do collision-model complexity and the number of robots affect the conflict scan benefit of \MRSPITE; (2) how does conflict scan reduction translate into planning time, and (3) how much pair-timestep fallback work and detailed sphere-pair work does \MRSPITE eliminate?

\subsection{Experimental Design}
To answer question (1), we isolate the conflict scan behavior by planning independent paths and recording performance of the first instance of the find-first-conflict operation on these paths (Sec.~\ref{sec:model_complexity}.) This is consistent with the first steps of many conflict-driven planners~\cite{sharon2015_cbs, solis2021_cbsmp, sim2025_stcbs, solis-arc-24}. To evaluate the impact of MR-SPITE as the complexity of the robot model and the number of robots change, we consider rigid body robots with increasing numbers of spheres in their collision geometry and consider team sizes of 4, 8, 16, and 32.  We use \VAMP as the underlying collision checker, a recent state-of-the-art fast validation method \cite{vamp_2024}. 
We evaluate the performance of \MRSPITE with \VAMP as the underlying validity checker against \VAMP by itself (no link filtering in either approach as these are single body robots) and report the conflict scan time over 100 random sets of paths (Figure~\ref{fig:mobile_sphere_count}).

To answer question (2), we evaluate the impact of MR-SPITE on a conflict-driven planner, extending the implementation of ARC provided in~\cite{motes2026_va_mrmp} to match the configuration described in Section~\ref{sec:arc-integration}.  To answer question (3), we measure the fraction of baseline pair-timestep candidates sent to the fallback and the sphere-pair work remaining relative to the \PRM+\VAMP baseline, together with the fraction of swept link-OBB pairs overlapping, on the same matched trials used in the planner experiments.
We use four different configurations of \MRSPITE with \VAMP: 
\MRSPITE or \MRSPITEH as the first collision scanning filter
with either VAMP or VAMP-H as a dense fallback. 
All \MRSPITE variants perform the same swept
per-link OBB tests. The \MRSPITEH variants additionally propagate the
resulting link mask to restrict configuration-level validation. 
We also include four ARC baselines without \MRSPITE, which are \RRTC or \PRM for initial paths with \VAMP or \VAMPH for conflict detection.

The first set of planner experiments uses mobile Fetch robots, represented by 111 collision spheres
distributed over 15 collision-bearing links. The Fetch joints are held
at a fixed nominal configuration while the mobile base moves in
$\mathrm{SE}(2)$ in the mobile crossing scene. This preserves the robot's complex multi-link
collision geometry while isolating the effect of team size in a mobile scene from
articulated-arm planning. The second set of planner experiments uses the 7-DoF Franka Emika Panda robot operating
around a cage as shown in Fig.~\ref{fig:panda_cage}. It is included to study the effects of the
conflict scanner on articulated robots.

All computed bounds are constructed using the same sphere-based collision model evaluated by the fallback scanner.
All \PRM-based variants use the same precomputed roadmaps and
swept-volume data. Roadmap and swept-bound construction are
performed offline and excluded from query planning time. 
We report scan time, which includes all of the inter-robot conflict scans performed by \ARC as it performs local repairs and updates the paths.
Planning time includes initial-path computation, conflict
scanning, local repair, and updating the repaired paths. Unless stated otherwise, we report the median
and interquartile range $[Q_1,Q_3]$. Scanner speedups are computed relative to \PRM + \VAMP.

Experiments were performed on a Linux workstation with an Intel Core
i7-14700F CPU and 62~GiB of memory.

\subsection{Scaling with Team Size and Model Complexity}\label{sec:model_complexity}
\begin{figure*}
    \centering
    \includegraphics[width=0.9\linewidth]{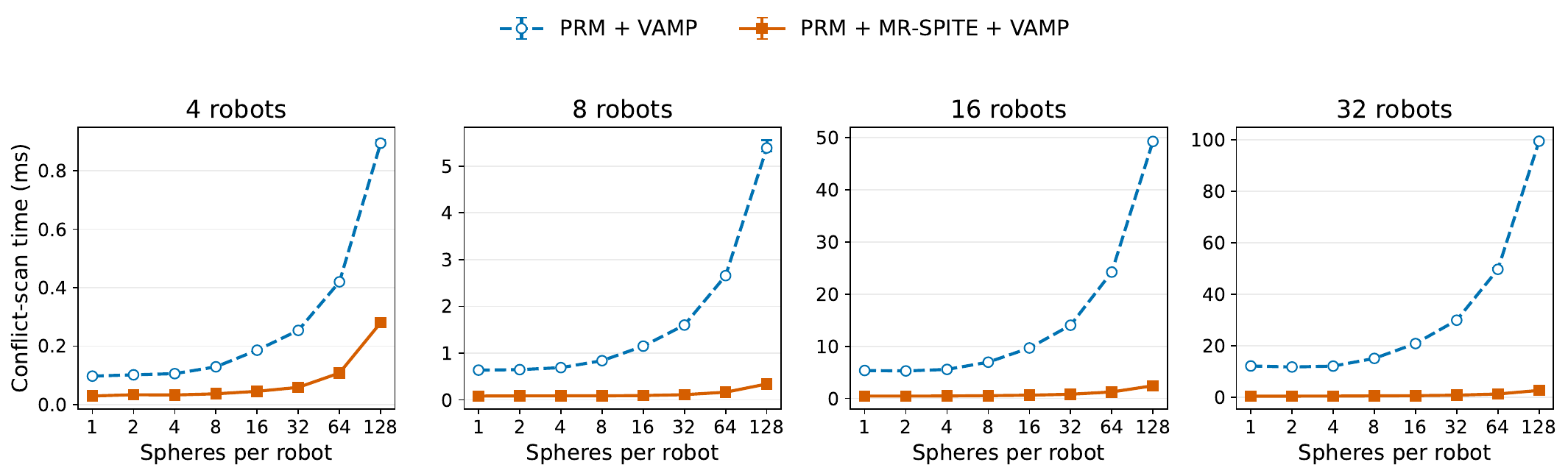}
    \caption{\small \textbf{Model Complexity Analysis}. The figure shows the effect of varying model complexity on conflict-scanner performance. The x-axis shows the number of spheres representing a spherical robot with an equal footprint across all counts. The tests were run  on 4, 8, 16, and 32 robots on the mobile crossing scene, on 100 sets of initial paths.}
    \label{fig:mobile_sphere_count}
    \vspace{-5mm}
\end{figure*}

Fig.~\ref{fig:mobile_sphere_count} shows how model complexity 
affects conflict scanner performance.
The experiment isolates conflict scanner cost by executing
one scan on synchronized PRM initial paths, before ARC performs any repair.
The results show a decrease in median conflict scan
time at every tested sphere count and team size. This includes the
one-sphere case, showing that interval scheduling and whole-motion
rejection can be useful even when an individual configuration-level
collision test is inexpensive.

These results indicate that \MRSPITE is particularly useful for robots
whose collision models require many geometric primitives.
Baseline VAMP scan time increases sharply with both team size and sphere count, reaching median times of 49.3 ms and 99.4 ms for 16 and 32 robots with 128 spheres per robot.
MR-SPITE grows much more slowly, requiring 2.43 ms and 2.68 ms in those same settings, 
corresponding to $20.3\times$  and $37.1\times$  median speedups in conflict scan time relative to VAMP.

For each interval pair, \MRSPITE requires one OBB check per link pair, 
independent of the number of underlying spheres. 
When these bounds are disjoint, it rejects the entire interval without
submitting its timesteps to detailed VAMP collision checking.
In contrast, the potential detailed sphere-pair work for a robot
pair grows with the product of their sphere counts.
Rejecting an interval therefore avoids dramatically more collision-checking work for geometrically
detailed robot models. The benefit also grows with team size because
more pairwise trajectories must be scanned.

\subsection{Planner Integration}\label{sec:planner_exp}

The top of Table~\ref{tab:results_all} reports results of running the
planner experiments with 4, 8, and 16 Fetch robots on the mobile crossing scenario.
Every displayed method succeeded on all tasks in its corresponding set.
With four robots, \VAMPH reduces the median conflict scan
time of \PRM{}+VAMP by 46\%, while \MRSPITEH{}+VAMP-H achieves a
79\% reduction. At this scale, however, supplemental-bound
construction is not fully amortized. Consequently, \PRM{}+VAMP-H,
rather than an \MRSPITE variant, obtains the lowest planning-time
median, reducing it by 27\% relative to \PRM{}+VAMP.

The end-to-end benefit of interval filtering increases with team size.
For eight robots, \MRSPITE{}+VAMP-H reduces the scan-time median by
81\% and the planning-time median by 34\% relative to \PRM{}+VAMP.
For sixteen robots, the corresponding reductions increase to 85\% and
57\%, respectively. Thus, as the number of trajectory pairs grows, the
conflict scan savings increasingly outweigh the cost of constructing
supplemental interval geometry.

The component variants show that the two filtering levels provide
independent benefits. \VAMPH alone provides paired scan speedups of
up to $1.88\times$, while \MRSPITE with standard \VAMP provides up to
a $3.36\times$ speedup. Propagating the surviving swept link pairs into
a restricted VAMP fallback with \MRSPITEH increases the speedup to as much as $5.29\times$.

At eight and sixteen robots, \MRSPITE{}+VAMP-H provides the lowest
scan and planning medians. Propagating the retained link mask into
\VAMPH increases the scan medians from 10.9 to 11.8~ms and from 71.5
to 76.4~ms, respectively. Although the retained-link restriction is
effective with standard VAMP, its additional mask and restricted
fallback overhead is not recovered when combined with VAMP-H in these
experiments.

The bottom of Table~\ref{tab:results_all} shows the results with the
articulated Panda manipulators. The four-robot experiment contains
30 tasks, while the eight-robot experiment contains 15 tasks. Every
displayed method succeeded on all tasks in its corresponding set.

For four Pandas, \MRSPITEH{}+VAMP obtains the lowest scan median,
reducing PRM{}+VAMP from 18.0 to 6.7~ms for a paired
$2.62\times$ speedup. Median planning time decreases from 43.4 to
38.0~ms. For eight Pandas, it reduces scan time from 257.2 to
112.4~ms, a paired $2.23\times$ speedup, and reduces planning time
from 741.9 to 690.0~ms.

These end-to-end reductions in planning time are
proportionally smaller than the scan reductions, in particular when compared to
the Fetch experiments, because initial-path search and local
repair account for a larger fraction of runtime in the constrained cage environment, and
supplemental \AABB construction adds a median of 24.4~ms for eight
robots. However, even with these additional costs, swept-interval
rejection and the link-restricted fallback improve
total planning performance for articulated,
high-dimensional robots as well as planar mobile systems.

The preferred fallback differs between the Fetch and Panda
experiments. VAMP-H represents each link group with a single enclosing
sphere. For the frozen Fetch model, this provides an effective
configuration-level filter, making \MRSPITE + \VAMPH the fastest
combination for larger teams. For the Panda, the enclosing spheres can
be loose approximations of its elongated links. The per-link swept
OBBs computed by \MRSPITE provide tighter geometric information, and
\MRSPITEH propagates their surviving link pairs into the VAMP
fallback. Consequently, \MRSPITEH{}+VAMP performs best in the Panda
experiments.

\RRTC{}+VAMP is included as a planner-level reference in both sets of experiments. 
Because \RRTC constructs different paths and invokes collision checking differently,
its runtime cannot be used to isolate the effect of the conflict
scanner. The paired PRM comparisons provide the controlled evaluation
of \MRSPITE on identical initial paths.

\subsection{Scanner-Work Reduction}

Table~\ref{tab:scanner-mechanism} explains the work reductions underlying
the observed scan-time improvements. We see that for all scenarios no more than 40\% of original fallback timestep pairs remain after \MRSPITE filtering, with as little as 5\% with 16 Fetch robots. In the Fetch scenarios, after the VAMP-H hierarchy is applied, only
0.6--1.3\% of the sphere-pair work remains.
Panda robots are more coupled in their interactions, but still detailed sphere-pair work falls
to around 10-20\%. Lastly, we observe that only 3.3--7.1\% of tested swept link-OBB pairs are retained as potentially overlapping. 
These measurements show that the runtime
improvements correspond to substantial reductions at both stages of the
scanner: fewer pair--timestep configurations reach the fallback checker, and the remaining calls require only a fraction of the baseline detailed geometric work.

\subsection{Correctness and Solution Quality}

Across all matched PRM trials, the scanner variants produced identical
initial paths, earliest-conflict sequences, repair
counts, makespans, and path sums. \MRSPITE therefore changes the
amount of collision-checking work performed without changing the
conflict decisions or resulting solutions of the underlying
discretized planner.

\begin{table*}
\vspace*{4pt}
\centering
\caption{Planner results. Scan, \AABB-construction, and 
planning times are median [Q1, Q3] in milliseconds.
Build is the one-time preprocessing time for roadmap construction and,
for MR-SPITE variants, swept-bound construction.
Speedups are paired medians relative to PRM+VAMP. }
\label{tab:results_all}
\scriptsize
\begin{tabular}{@{}lllrllllll@{}}
\toprule
Scene & $\#$ & Method & Scanner & Success & Build (s) & Scan [IQR] (ms) & \AABB [IQR] (ms) & Plan [IQR] (ms) & Speedup (x) \\
\midrule
Fetch Cross & 4 & RRTC & VAMP & 30/30 & -- & 8.6 [7.0, 9.9] & -- & 29.1 [25.6, 34.2] & -- \\
 &  & RRTC & VAMP-H & 30/30 & -- & 4.5 [4.2, 5.5] & -- & 25.9 [24.5, 31.3] & -- \\
 &  & PRM & VAMP & 30/30 & 0.22 & 8.5 [7.0, 11.3] & -- & 20.0 [15.5, 23.2] & 1.00 \\
 &  & PRM & VAMP-H & 30/30 & 0.22 & 4.6 [4.4, 5.1] & -- & \textbf{14.7 [12.7, 17.7]} & 1.88 \\
 &  & PRM & MR-SPITE + VAMP & 30/30 & 0.95 & 4.8 [3.6, 7.1] & 4.2 [3.7, 5.7] & 22.4 [17.9, 26.9] & 1.77 \\
 &  & PRM & MR-SPITE-H + VAMP & 30/30 & 0.95 & 2.9 [2.4, 3.6] & 4.2 [3.8, 5.6] & 18.7 [16.4, 23.2] & 2.94 \\
 &  & PRM & MR-SPITE + VAMP-H & 30/30 & 0.95 & 1.9 [1.6, 2.3] & 4.1 [3.9, 5.6] & 17.3 [16.0, 21.4] & 4.52 \\
 &  & PRM & MR-SPITE-H + VAMP-H & 30/30 & 0.95 & \textbf{1.8 [1.6, 2.3]} & 4.5 [3.8, 5.6] & 17.8 [16.1, 21.4] & 4.67 \\
\midrule
 & 8 & RRTC & VAMP & 30/30 & -- & 53.4 [50.0, 61.7] & -- & 116.3 [108.0, 123.8] & -- \\
 &  & RRTC & VAMP-H & 30/30 & -- & 35.9 [34.6, 41.5] & -- & 100.3 [93.1, 106.5] & -- \\
 &  & PRM & VAMP & 30/30 & 0.20 & 58.8 [52.4, 65.8] & -- & 93.0 [82.0, 102.8] & 1.00 \\
 &  & PRM & VAMP-H & 30/30 & 0.20 & 40.9 [37.5, 44.2] & -- & 72.7 [66.7, 82.0] & 1.45 \\
 &  & PRM & MR-SPITE + VAMP & 30/30 & 0.90 & 20.0 [15.9, 24.2] & 13.1 [11.1, 15.9] & 71.5 [63.4, 84.0] & 2.99 \\
 &  & PRM & MR-SPITE-H + VAMP & 30/30 & 0.90 & 15.3 [12.7, 17.2] & 13.2 [10.9, 15.8] & 66.6 [60.4, 77.2] & 3.90 \\
 &  & PRM & MR-SPITE + VAMP-H & 30/30 & 0.90 & \textbf{10.9 [9.3, 12.5]} & 13.0 [11.0, 15.9] & \textbf{61.5 [55.5, 72.7]} & 5.59 \\
 &  & PRM & MR-SPITE-H + VAMP-H & 30/30 & 0.90 & 11.8 [10.0, 13.9] & 12.8 [10.8, 15.8] & 62.7 [56.2, 74.0] & 5.02 \\
\midrule
 & 16 & RRTC & VAMP & 30/30 & -- & 372.3 [356.8, 403.2] & -- & 580.9 [556.4, 617.8] & -- \\
 &  & RRTC & VAMP-H & 30/30 & -- & 282.4 [270.2, 301.8] & -- & 491.1 [474.0, 517.7] & -- \\
 &  & PRM & VAMP & 30/30 & 0.19 & 490.9 [475.5, 574.0] & -- & 641.0 [596.8, 732.0] & 1.00 \\
 &  & PRM & VAMP-H & 30/30 & 0.19 & 338.0 [317.4, 400.6] & -- & 491.0 [439.6, 565.7] & 1.48 \\
 &  & PRM & MR-SPITE + VAMP & 30/30 & 0.87 & 146.2 [130.9, 174.4] & 33.4 [28.2, 44.4] & 354.5 [302.6, 413.5] & 3.36 \\
 &  & PRM & MR-SPITE-H + VAMP & 30/30 & 0.87 & 96.2 [83.0, 124.3] & 33.3 [29.1, 44.2] & 302.9 [258.0, 357.3] & 5.29 \\
 &  & PRM & MR-SPITE + VAMP-H & 30/30 & 0.87 & \textbf{71.5 [61.7, 91.5]} & 32.9 [28.5, 44.9] & \textbf{276.8 [233.7, 332.5]} & 7.18 \\
 &  & PRM & MR-SPITE-H + VAMP-H & 30/30 & 0.87 & 76.4 [67.2, 103.7] & 36.4 [29.3, 45.8] & 283.0 [241.0, 351.0] & 6.54 \\
\midrule
Panda Cage & 4 & RRTC & VAMP & 30/30 & -- & 15.4 [9.6, 21.4] & -- & 49.1 [24.5, 115.2] & -- \\
 &  & RRTC & VAMP-H & 30/30 & -- & 18.5 [11.7, 21.9] & -- & 52.3 [27.8, 121.0] & -- \\
 &  & PRM & VAMP & 30/30 & 0.58 & 18.0 [10.5, 29.9] & -- & 43.4 [16.9, 129.3] & 1.00 \\
 &  & PRM & VAMP-H & 30/30 & 0.58 & 16.8 [8.9, 22.0] & -- & 39.6 [16.5, 116.2] & 1.13 \\
 &  & PRM & MR-SPITE + VAMP & 30/30 & 2.58 & 13.7 [7.2, 23.8] & 2.7 [0.8, 4.5] & 44.8 [16.4, 135.9] & 1.32 \\
 &  & PRM & MR-SPITE-H + VAMP & 30/30 & 2.58 & \textbf{6.7 [3.4, 14.3]} & 2.7 [0.9, 4.6] & \textbf{38.0 [11.9, 117.1]} & 2.62 \\
 &  & PRM & MR-SPITE + VAMP-H & 30/30 & 2.58 & 12.1 [6.3, 18.5] & 2.7 [0.8, 4.5] & 43.2 [15.6, 125.5] & 1.53 \\
 &  & PRM & MR-SPITE-H + VAMP-H & 30/30 & 2.58 & 19.9 [6.6, 29.6] & 2.7 [0.8, 4.6] & 50.3 [16.2, 147.2] & 1.01 \\
\midrule
 & 8 & RRTC & VAMP & 15/15 & -- & 224.9 [140.8, 272.0] & -- & 778.9 [484.9, 1302.9] & -- \\
 &  & RRTC & VAMP-H & 15/15 & -- & 203.7 [135.1, 251.1] & -- & 778.5 [471.7, 1294.1] & -- \\
 &  & PRM & VAMP & 15/15 & 1.53 & 257.2 [180.9, 456.8] & -- & 741.9 [457.6, 1420.1] & 1.00 \\
 &  & PRM & VAMP-H & 15/15 & 1.53 & 226.2 [169.2, 390.9] & -- & 731.5 [434.6, 1335.3] & 1.13 \\
 &  & PRM & MR-SPITE + VAMP & 15/15 & 19.61 & 161.5 [103.5, 312.5] & 24.2 [12.7, 43.4] & 703.3 [379.2, 1344.9] & 1.49 \\
 &  & PRM & MR-SPITE-H + VAMP & 15/15 & 19.61 & \textbf{112.4 [76.6, 211.7]} & 24.4 [12.7, 41.6] & \textbf{690.0 [343.7, 1233.2]} & 2.23 \\
 &  & PRM & MR-SPITE + VAMP-H & 15/15 & 19.61 & 135.6 [99.5, 261.1] & 24.6 [12.8, 41.4] & 698.7 [371.5, 1273.7] & 1.79 \\
 &  & PRM & MR-SPITE-H + VAMP-H & 15/15 & 19.61 & 205.0 [126.3, 394.6] & 24.3 [12.8, 43.8] & 736.4 [407.0, 1397.0] & 1.30 \\
\bottomrule
\end{tabular}
\vspace{-2mm}
\end{table*}

\begin{table}
\centering
\caption{Scanner work on matched planner trials (median [Q1, Q3]).
Values are percentages relative to PRM+VAMP, except swept link pairs overlapping, which is relative to tested link-OBB pairs.}
\label{tab:scanner-mechanism}
\scriptsize
\begin{tabular}{@{}lrrr@{}}
\toprule
Robots &
\makecell{Pair--timesteps\\to fallback (\%)} &
\makecell{Sphere-pairs\\ checked (\%)} &
\makecell{Swept link\\pairs overlapping (\%)} \\
\midrule

\multicolumn{4}{@{}l}{\textit{Fetch perpendicular:
MR-SPITE + VAMP-H}} \\
4  & 21.9 [20.4, 26.2] & 0.6 [0.4, 1.2] & 5.7 [5.1, 6.2] \\
8  & 10.3 [9.5, 10.9]  & 0.7 [0.5, 0.7] & 4.4 [3.9, 4.6] \\
16 & 5.0 [4.8, 5.3]    & 1.3 [1.2, 1.4] & 3.3 [3.1, 3.5] \\

\midrule
\multicolumn{4}{@{}l}{\textit{Panda cage:
MR-SPITE-H + VAMP}} \\
4 & 39.4 [26.4, 49.2] & 20.7 [15.7, 26.1] & 7.1 [5.1, 9.9] \\
8 & 25.5 [20.3, 29.5] & 12.8 [10.5, 14.9] & 5.9 [4.7, 7.7] \\

\bottomrule
\end{tabular}
\vspace{-4mm}
\end{table}

\section{Conclusion}

We introduced \MRSPITE, a conservative motion-interval filter for
accelerating find-first-conflict scans over synchronized multi-robot
paths. \MRSPITE partitions paths according to their motion provenance,
reuses precomputed swept \OBB{}s for PRM edges, and constructs
supplemental \AABB{}s for query connectors, goal holds, and local
repairs. Its interval scheduler uses these bounds to certify complete
temporal windows as conflict-free before invoking a detailed
configuration-level collision checker. This allows \MRSPITE to
complement existing accelerators such as \VAMP without changing the
temporal discretization or conflict semantics of the underlying
planner.

Integrated into \ARC, \MRSPITE consistently reduced conflict scanning
cost across the tested team sizes and collision-model complexities.
For 16 Fetch robots, \MRSPITE{}+VAMP-H achieved a paired
$7.18\times$ scan speedup over \PRM{}+VAMP and reduced median planning
time by 57\%. Propagating surviving swept link pairs into a restricted
VAMP fallback provided up to a paired $2.62\times$ scan speedup for
the articulated Panda manipulators. The sphere-count study showed that \MRSPITE
reduced scan time even for simple collision models, while its
planning time benefit increased as detailed collision checking became
more expensive.
The results also demonstrate that additional hierarchy is not
universally beneficial. Although restricting standard VAMP using the surviving link mask improves performance,
combining it with \VAMPH introduces additional masked-fallback
overhead and does not outperform the simpler \MRSPITE{}+VAMP-H
combination on the larger Fetch problems. Furthermore, \MRSPITE's
online bound-construction cost may outweigh scan savings for
small, inexpensive problems. Future work will investigate tighter and cheaper supplemental bounds and integration with other conflict-driven planners, including CBS-based methods.

\section*{Acknowledgment}
\vspace{-2mm}

OpenAI Codex was used to generate or refactor portions of the collision-detection code and scripts used to run experiments and generate plots. 
All AI-generated code was reviewed, tested, and modified by the authors.

\bibliographystyle{IEEEtran}
\bibliography{references}

\end{document}